%% file: main.tex
\documentclass{article}

\usepackage[preprint,nonatbib]{neurips_2019}

\usepackage{comment} 
\usepackage{xcolor}
\usepackage[utf8]{inputenc} 
\usepackage[ruled,vlined]{algorithm2e} 
\usepackage{graphicx}

\usepackage{caption}
\usepackage{subcaption}
\usepackage{amsmath} 
\usepackage{dsfont} 

\usepackage{ amssymb } 
\usepackage{multirow}

\usepackage{hyperref} 
\usepackage[normalem]{ulem}

\usepackage[T1]{fontenc}    
\usepackage{hyperref}       
\usepackage{url}            
\usepackage{booktabs}       
\usepackage{amsfonts}       
\usepackage{nicefrac}       
\usepackage{microtype}      

\newcommand{\overbar}[1]{\mkern 1.5mu\overline{\mkern-1.5mu#1\mkern-1.5mu}\mkern 1.5mu}

\DeclareMathOperator*{\argmin}{arg\,min}

\title{Sparsely Grouped Input Variables \\ for Neural Networks}

\author{
Beibin Li$^{\star \dagger}$
\And
Nicholas Nuechterlein$^{\star}$ \And
Erin Barney$^\dagger$ \And
Caitlin Hudac$\dagger \ddagger$\AND
Pamela Ventola$^\mathsection$\And
Linda Shapiro$^\star$\And
Frederick Shic$^{\star \dagger \ddagger}$\And~
\vspace{-6mm} \\
$^{\star}$ Paul G. Allen School of Computer Science and Engineering, University of Washington, Seattle, WA ~\\
$^{\dagger}$ Seattle Children's Research Institute, Seattle, WA ~\\
$^{\ddagger}$ School of Medicine, University of Washington, Seattle, WA ~\\
$^{\mathsection}$ Yale Child Study Center, School of Medicine, Yale University, New Haven, CT
}

\begin{document}

\maketitle

\begin{abstract}
In genomic analysis, biomarker discovery, image recognition, and other systems involving machine learning, input variables can often be organized into different groups by their source or semantic category. Eliminating some groups of variables can expedite the process of data acquisition
and avoid over-fitting. Researchers have used the group lasso to ensure group sparsity in linear models and have extended it to create compact neural networks in meta-learning. Different from previous studies, we use multi-layer non-linear neural networks to find sparse groups for input variables.
We propose a new loss function to regularize parameters for grouped input variables, design a new optimization algorithm for this loss function, and test these methods in three real-world settings. 
We achieve group sparsity for three datasets, maintaining satisfying results while excluding one nucleotide position from an RNA splicing experiment, excluding 89.9\% of stimuli from an eye-tracking experiment, and excluding 60\% of image rows from an experiment on the MNIST dataset.
\end{abstract}

\section{Introduction}

In machine learning pipelines, input variables can often be organized into distinct groups. For instance, a medical diagnosis system usually requires results from several biological tests (e.g., blood tests, MRI scans, and genetic screenings, etc.). We can organize these results into groups of features, where each group is comprised of all features derived from a particular test.

Machine learning systems are usually more robust when provided with input data from separate, complementary sources, which can supply additional context and whose redundancy can reinforce model stability, as in our medical diagnosis system example. However, there are two fundamental reasons to remove highly-correlated groups of features: to reduce data acquisition cost and to avoid over-fitting. We say a group is sparse when all model parameters associated with the input variables in that group are zero. Sparse groups are said to have been removed by the model. We will say that a model has achieved sparsity when it has removed groups of variables during training.

In many real-world machine learning problems, obtaining additional input data is extremely expensive. While general feature selection methods, such as the standard lasso, can prune features to help mitigate over-fitting, they do not take into account how features can be grouped together, such as by acquisition source or semantic category.  In particular, if features are selected at the group level, highly-correlated groups of features may be removed and subsequent feature acquisition need not collect data from the sources of these discarded groups, which may substantially reduce data acquisition cost, promote model concision, and reduce the model’s vulnerability to over-fitting. 


Ensuring group sparsity is the process of removing redundant groups from a model's input data, which is closely related to feature selection in machine learning. Exploring group sparsity has many practical applications for machine learning systems.
For instance, in eye-tracking experiments for children with autism spectrum disorder (ASD) shown in Section \ref{sec:experiment_asd}, the machine learning model’s performance does not decrease even as groups of input features are removed, likely because many of the different feature groups encode similar information. This is practically advantageous because shorter experiments are more tolerable for children and  also allow for the collection of data from a broader set of participants.

Researchers use many existing techniques to partially solve this problem. When the number of groups, $k$, is small, people can use a naive brute-force approach to train $O(\binom{k}{m})$ models to select the best $m$ sub-groups or to train $O(2^k)$ models to find the optimal combination of groups. However this method becomes intractable for larger and more complex data.
People can also perform multi-stage machine learning (e.g., \cite{yan2008multimodal,lisowska2018joint}) in which data reduction techniques such as feature selection, PCA, LDA, or autoencoders, are used to extract information for each group in the first stage before feeding the results to another machine learning model.  
This strategy is straightforward and interpretable, but the characteristics and assumptions of the architecture can limit discovery of critical feature interactions. In addition, group-based variable selection techniques such as Yuan and Lin's group lasso \cite{yuan2006model}  can be trained in polynomial time and allow for linear combination of variables from different groups. However, these methods are typically linear in nature and could not generalize to nonlinear problems.

\begin{figure}[t]
  \centering
  \includegraphics[width=0.3\textwidth]{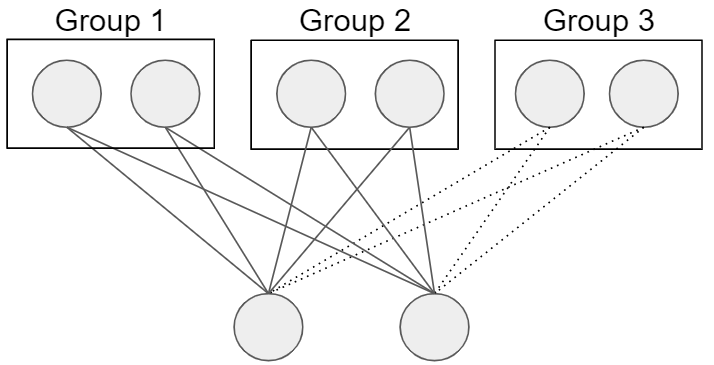}
  \includegraphics[width=0.6\textwidth]{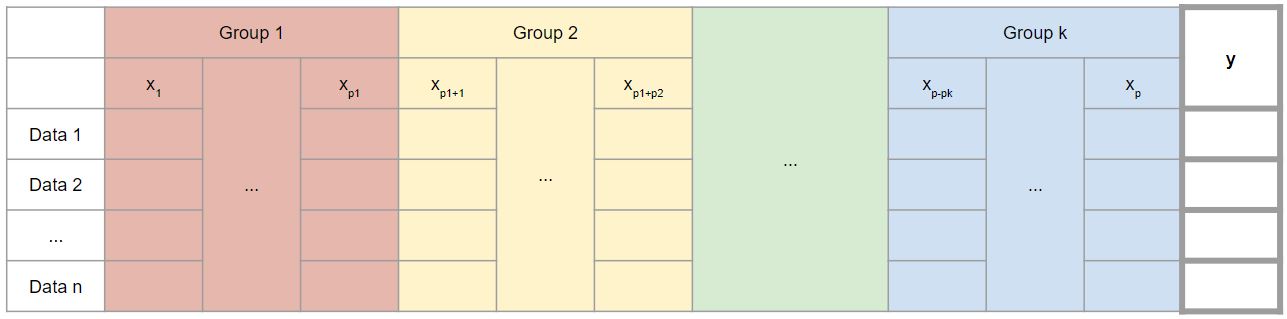}
  \caption{
  \textit{Left:} Sparsity for grouped input variables. The top row is the input layer; the bottom row is the first hidden layer; dashed lines stand for connections that can be removed to satisfy the sparsity constraint if group $3$ is redundant (and thus group $3$ could subsequently be removed in the final ML pipeline).  
  \textit{Right:} Sample data format: we have $n$ data points and $k$ groups of variables. 
  }
  \label{fig:sparse}
\end{figure}

In the last decade, neural networks (NNs) have gained popularity and tend to outperform linear models when the data is not linearly separable.
We would like to integrate the group lasso with non-linear multi-layer NNs so that they can also satisfy group sparsity constraints.
In NNs, the sparsity for a given group must be ensured for all the neurons in the first hidden layer, as shown in Figure \ref{fig:sparse}, not just the single output in the group lasso.
Moreover, popular existing optimization algorithms, such as stochastic gradient descent (SGD), usually do not satisfy the sparsity requirements even if they can converge given an infinite amount of time. This motivates the development of a new method to train neural networks adapted for satisfying sparse group constraints.

In this paper, we (1) propose a new loss function for group regularization in fully-connected layers, 
(2) create a novel optimization algorithm to learn suitable sparse network parameters, 
and (3)  propose SGIN  (\textbf{S}parsely \textbf{G}rouped variables \textbf{I}n \textbf{N}eural Network) model to ensure group sparsity in NNs.
SGIN uses the same representation and inference systems as NNs so that existing deep learning tools can be applied directly to use SGIN.
Nevertheless, the learning phase of SGIN is significantly different from traditional NNs: training SGIN requires stochastic, blockwise, and coordinal optimization methods, which is significantly different from training traditional NNs.

\section{Related Work}
\label{sec:related}

Sparsity has been studied in machine learning models for decades, and the lasso is perhaps the most famous example. As shown in Equation \ref{eq:lasso}, the standard lasso is a linear model that utilizes a $\lambda$-weighted $l$-1 regularization to enforce sparsity in the model's input variables \cite{tibshirani1996regression}. The group lasso introduces the notion of groups to extend the standard lasso. Let $X \in \mathbb{R}^{n\times p}$ be the matrix of $n$ samples and $p$ features, $y \in \mathbb{R}^n$ be the vector of $n$ labels, and $\beta^* \in \mathbb{R}^p$ be the vector of optimal model parameters.

Formally, the group lasso partitions the $p$ features into $k$ sets, each of which we will refer to as a group $X_i$ for $i \in \{1, \ldots, k\}$. We will denote the cardinality of each $X_i$ as $p_i$ so that we can rewrite the matrix $X = \{X_1, X_2, ..., X_k\}$ and parameter vector $\beta = \{\beta_1, \beta_2, ..., \beta_k\}$, where $X_i \in \mathbb{R}^{n \times p_i}$ represents $p_i$ features  and $\beta_i \in \mathbb{R}^{p_i}$ represents the associated parameters for group $X_i$. 
The  norm for regularization is not squared in the group lasso; 
otherwise, it becomes weighted ridge regression.

\begin{align}
    \text{Lasso:~~~} &  \beta^* = \argmin_\beta || y - X \beta ||^2_2 + \lambda || \beta ||_1 \label{eq:lasso} \\
    \text{Group lasso:~~~} & \beta^* = \argmin_\beta || y - \sum\limits_{i=1}^k X_i \beta_i ||^2_2 + \lambda \sum\limits_{i=1}^k \sqrt{p_i} || \beta_i ||_2
    \label{eq:group_lasso}
\end{align}

Thousands of studies have extended the group lasso to fulfill different objectives, including
bi-level sparsity \cite{huang2009group,simon2013sparse,frecon2018bilevel},
non-linear logistic regression \cite{meier2008group},
tree-structures \cite{kim2010tree},
overlapping groups \cite{jacob2009group,yuan2011efficient}, 
hierarchical interaction \cite{bien2013lasso}, 
Bayesian variable selection \cite{xu2015bayesian}, 
graphical models \cite{friedman2008sparse},
among many other interesting variations \cite{huang2012selective}. 
We compare our results with \cite{yuan2006model,meier2008group} in Section \ref{sec:experiments}, where \cite{yuan2006model} is the original group lasso and \cite{meier2008group} is a robust method that combines advantages from log-likelihood and group lasso.
Besides lasso-based models, 
sparse coding approach \cite{olshausen1997sparse},  
sparse matrix decomposition \cite{zou2006sparse,witten2009penalized,chandrasekaran2011rank}, sparse Bayesian learning \cite{tipping2001sparse}, sparse deep belief network \cite{lee2008sparse}, 
and many other works in sparsity also inspire us to extend feature sparsity in grouped nonlinear setting.

More recently, researchers have applied  $l$-1 regularization to create compact and efficient NN structures \cite{alvarez2016learning,scardapane2017group,yoon2017combined,alford2018pruned} by pruning  neurons. The concept is even extended to Convolutional Neural Network in Structured Sparsity Learning \cite{wen2016learning}. 
Beyond the $l$-1 regularization, dynamic routing \cite{sabour2017dynamic}, EM routing \cite{hinton2018matrix}, dropout \cite{srivastava2014dropout}, group normalization \cite{wu2018group}, self normalization \cite{klambauer2017self}, grouped linear transformation \cite{mehta2018pyramidal}, and many other methods have been developed to augment and improve NNs. These innovative methods improve convergence and performance in neural networks, but they do not solve the group sparsity problem because the assumptions and goals are different in a group setting. Nevertheless, these methods inspire us and helped us to create a robust and efficient algorithm.

Our primary goal is to discard groups of input variables under the assumption that there are no hierarchical or overlapping groups (i.e., each variable belongs to exactly one feature group).
We will extend group $l$-1 regularization to multi-layer non-linear NNs and ensure sparsity in the first hidden fully-connected layer. Different from \cite{yuan2006model,meier2008group} and related studies whose primary goal is regularization, we focus primarily on removing groups of input variables in order to reduced the information a machine learning system requires during inference and to ease the expense of future data collection.

\section{Methods}
\label{sec:methods}

To generalize our method to all problems, we let $J$ be the element-wise loss and $\phi$ be an arbitrary loss function in a machine learning system.
We define the optimization goal for $\beta$, the vector of all parameters in the NN model $f$, and the regularization term $\tau$ as follows.

\begin{align}
   \text{Optimal Set of Parameters:~~~}  & \beta^*  =
   \argmin_\beta  J(\beta) + \lambda \tau =
   \argmin_\beta  \phi(f(X), y) + \lambda \tau
   \label{eq:gsnn_loss}\\
   \text{Regularization Term:~~~}  & \tau = \sum\limits_{i=1}^k \tau_i = 
   \sum\limits_{i=1}^k \sqrt{p_i}  || \theta_{i} ||_2 \label{eq:gsnn_reg}
\end{align}
Here $\theta \in \mathbb{R}^{p \times q}$ is the matrix of parameters in the first layer of the NN, $q$ is the number of neurons in the first hidden layer, and $\theta \subset \beta$. 
We let $\theta_i \in \mathbb{R}^{p_i \times q}$ be the parameters between the input variables in group $i$ and all neurons in the first hidden layer.

The  $\tau_i$ is the Frobenius norm of the $\theta_i$ matrix rather than the norm of the $\beta_i$ vector in Equation \ref{eq:group_lasso}.
When there is no hidden layer and $q = 1$, then the regularization term $\tau$ becomes equivalent to the group lasso. 
When $k$ and $p$ are equivalent, each group is comprised of one $q$-dimensional feature vector, and $\tau$ reduces to the regularization of the first hidden layer in \cite{scardapane2017group}, an effective method to prune neurons from NNs.
Here, we only apply the regularization for the first fully-connected layer because only the parameters for the input layer should be pruned in our setting

Applying gradient descent (GD) or SGD directly to Equation \ref{eq:gsnn_loss} does not guarantee group sparsity for the NN, because SGD does not guarantee convergence if the loss is not differentiable. However, the regularization for each group $||\theta_i||_2$ is not differentiable at the origin, which is preferred by our loss function to sparse groups.
Using coordinate descent, blockwise coordinate descent (BCD), or graphical lasso algorithm is not feasible either because these methods can only optimize parameters in the first hidden layer. 
Instead, we combine these optimization methods to create the new optimization algorithm shown in Algorithm \ref{algo:sbcgd}.

\begin{algorithm}[h]
\SetAlgoLined
\LinesNumbered
\KwIn{Neural Network $f$, regularization term $\lambda$, loss function $\phi$, Training data $X$, training labels $y$}
\KwResult{Optimize the parameters $\mathbf{W}$ in the neural network $f$ for one epoch}
 Randomly shuffle the groups\;
\ForEach{Group $i \in [1,\ldots k]$}{ \label{alg:line_loop_begin}
   \lIf{group $i$ is sparse in $f$}{ continue } \label{alg:line_continue}
   arr = \{\}\;
   \ForEach{mini-batch $(X', y') \subset (X, y)$}{ \label{alg:line_batch_begin}
       $loss_1 = \phi(f(X'), y')$\; \label{alg:line_loss_1}
       $loss_2 = \phi(f_{\bar{i}}(\overbar{X'_i}), y')$\;
       arr.push(max($loss_2 - loss_1$, 0))\;
       Optimize $\Theta \cup \theta_i$ 
       according to $loss_1 + \lambda \tau_i$  by using Stochastic Gradient Descent\; \label{alg:line_sgd}
   } \label{alg:line_batch_end}
   \If{mean(arr) $< \lambda$}{ \label{alg:line_mean_arr}
       Set $\theta_i = 0$, and make group $i$ sparse in $f$\;
   }
}
\caption{Stochastic Blockwise Coordinated Gradient Descent}\label{algo:sbcgd}
\end{algorithm}

We denote $X' \in \mathbb{R}^{b\times p}$ and $y' \in \mathbb{R}^{b}$ as a subset (i.e., mini-batch) of the training set $X, y$, where $b$ is the batch size. 
We let $\Theta$ be all parameters in the NN except the parameters between the input and first hidden layer. That is, $\theta \cup \Theta = \beta$, $\theta \cap \Theta = \emptyset$.

Consider the case where group $X_i$ is excluded.
Let $\overbar{X_i} = \{X_1, X_2, ..., X_{i-1}, \mathbf{0}_{n, p_i}, X_{i+1}, ..., X_k\}$ 
represent the data where all features belonging to group $X_i$ are set to zero, and 
$f_{\bar{i}}$ is a copy of our NN $f$ where the connections between variables in group $X_i$ and all neurons in the first hidden layer are removed. 
Here, $\theta_{\bar{i}}$ represents the parameters in the first hidden-layer except for those associated with group $X_i$.

The back-propagation step at line \ref{alg:line_sgd} uses SGD and can help optimize all relevant parameters $\theta_{\bar{i}}$ in the first layer of $f$, which makes the SGD algorithm compatible with the BCD.
Define $f_{\beta_{\bar{i}}}$ as the neural network where parameters for group $i$ is sparse, we have the following proposition to guarantee line \ref{alg:line_mean_arr} to be a good approximation for coordinate convergence, where $mean(arr)$ is the sample-wise average for the difference of losses. 
If the parameters for group $i$ is near the non-differentiable point at line \ref{alg:line_mean_arr}, we can discard this group.
Another variant of this algorithm, which provides a sufficient and necessary condition for coordinate convergence, is in the Appendix for comparison, but Algorithm \ref{algo:sbcgd} is better in our observation.

\begin{align}
\text{\textbf{Proposition: }} & 
\mathbb{E}[\phi(f_{\beta_{\bar{i}}} (X_{\bar{i}}), y)]  -   \mathbb{E}[\phi(f_\beta(X), y)] < \lambda \tau_i
\Longleftrightarrow J( \beta_{\bar{i}}) < J(\beta)
\label{eq:proposition}
\end{align}

An alternative method would be to store the gradient for $\theta_{\bar{i}}$ during each mini-batch optimization and then perform a one-time update for $\theta_{\bar{i}}$ after line \ref{alg:line_batch_end}.
However, this approach is not crucial when $k$ or the number of training epochs is large enough, so we skip this step for mathematical succinctness and engineering convenience.
During optimization for $\theta_i$, the interactions among different groups can still be ensured in other hidden layers, and we will show this with a concise simulation for SGIN to learn a 8-way XOR (parity) operation in the Appendix; however, we admit this could limit the interactions in the first layer.
 
Another decision we make is once we make a group sparse, we will never visit it again, as shown in line \ref{alg:line_continue}.  This decision guarantees stability across different epochs.
The algorithm performs $O(\frac{n k}{b})$ back-propagation steps during each epoch. Even if it is $k$ times slower than the traditional SGD algorithm in each epoch, the number of epochs needed for our algorithm is smaller than the one needed for SGD,
which results in similar computation time for traditional SGD and Algorithm \ref{algo:sbcgd}.
We name neural network models trained with loss \ref{eq:gsnn_loss} and optimization Algorithm \ref{algo:sbcgd} SGIN.

\section{Empirical Experiments}
\label{sec:experiments}

In this section, we examine SGIN on three datasets.
For an ablation study, we compare SGIN with a traditional NN and ``NN + $\tau$", where ``NN + $\tau$" is a NN trained with the regularization term $\tau$ (equations  \ref{eq:gsnn_loss} and \ref{eq:gsnn_reg}) by SGD. 
We use the same neural network structure for SGIN, the NN, and the ``NN + $\tau$" in each of the following experiments.
We also compare our methods with the lasso, the group lasso, and the group lasso for logistic regression (GLLR) in the first two experiments.
We use $\mathds{1}_{\hat{y} > 0.5}$ to represent positive predictions in binary classification problems. 
More details are included in the Appendix.

For simplicity, we try 1-hidden layer, 2-hidden layer, and 3-hidden layer NN before the experiments as shown in Table \ref{tab:nn_structures}, and then choose the one with fastest training convergence. 
We do not test deeper models because model performance is not the ultimate goal of this study. 

\begin{table}[h!]
\centering
\begin{tabular}{l|ccc}
\toprule
               Structure & RNA Splicing              & Eye-tracking for ASD     & MNIST                        \\
               \midrule
1-hidden layer & {[}30{]}                  & {[}500{]}                & {[}100{]}                    \\
2-hidden layer & {[}30, 10{]}              & \textbf{{[}3000, 500{]}} & {[}1000, 100{]}              \\
3-hidden layer & \textbf{{[}30, 20, 10{]}} & {[}3000, 1000, 500{]}    & \textbf{{[}1000, 100, 50{]}} \\
    \bottomrule
\end{tabular}
\caption{The NN structures we tried for each experiment. Each numbers in an array represent the number of neurons in the corresponding hidden layers. Note that the input layers and output layers are not included in this table. The bold structure is the final structure used in experiments.}
\label{tab:nn_structures}
\end{table}

\subsection{Nucleotides in RNA Splicing}
\label{sec:experiment_rna}

In our first experiment, we attempt to discriminate real human 5' (donor) splice sites from decoys in a public  dataset \cite{yeo2004maximum}. This is a traditional binary classification problem in RNA splicing.
There are 8451 true human donor sites and 179438 false human donor sites in the training set and 4208 true human donor sites and 89717 false donor sites in the test set. 
Each data sample contains a sequence of seven nucleotides, which we represent with $7 \times 4 = 28$ features using one-hot vector encoding. We divide these features into seven groups according to nucleotide position.
Yeo and Burge \cite{yeo2004maximum} collected this dataset and achieved 0.6589 Maximum Correlation Coefficient (Max CC) using a  maximum entropy model which  outperformed all previous probabilistic models. 
Meier,  van de Geer, and Bühlmann created the GLLR and also achieved comparable results \cite{meier2008group}. 
Both studies give insight into the relative importance of dependencies in the RNA sequences; we will push further and investigate whether all seven nucleotide positions are necessary to distinguish real and decoy 5' splices sites.
We do not use prior knowledge in this experiment and leave the models to learn the interaction between nucleotides by themselves. 
Similar to \cite{yeo2004maximum,meier2008group}, we use Max CC to evaluate testing results.

\begin{table}[h]
\centering
    \begin{tabular}[width=0.9\textwidth]{cc|cccc}
    \hline
    \toprule
    \multirow{ 2}{*}{Sampling Method}  &   \multirow{ 2}{*}{Model}                          &     \multicolumn{4}{c}{Number of Sparse Groups} \\
     & & 0            & 1  & 2 & 3  \\
    \midrule
    \multirow{5}{*}{Down-sampling}       &   SGIN           & 0.6560 & 0.6411 & 0.5578 & \textbf{0.5507}               \\
                                         & NN + $\tau$    & 0.6270       & - & - & -            \\
                                         & NN            & 0.6225             & - & - & -      \\
                                         & Lasso & 0.6148 & 0.5627 & 0.5180 & 0.4782                                        \\ 
                                         & GLLR with \cite{meier2008group}  & 0.6208  & 0.5655 & 0.5599 & 0.5190             \\ 
    \hline
    \multirow{4}{*}{All}       & SGIN           & \textbf{0.6689} & \textbf{0.6432} & \textbf{0.5950} & 0.5394                 \\
                                         & NN + $\tau$    & 0.6657      & - & - & -           \\
                                         & NN             & 0.6646       & - & - & -            \\
    \bottomrule
     \end{tabular}
\caption{Testing Max CC for RNA Splicing:
SGIN outperforms all previous methods. 
When more than 4 groups are made sparse (out of the 7 groups), the Max CCs for all models drop below 0.5 and hence are not included in the table.
}
\label{tab:rna_rst}
\end{table}

To compare with \cite{meier2008group}, we randomly sample 5610 real and 5610 decoy examples from the training set without replacement and use the rest of original training set as a validation set. We then train SGIN with different $\lambda$ to obtain different group sparsity.
We use the R package provided by \cite{meier2008group}, who achieved 0.65 Max CC with three-way and lower order interactions extracted from the dataset. 
In our experiment, we exclude manual interaction extraction for fair comparison, and only managed to reach 0.62 Max CC with the GLLR.
Nevertheless, SGIN outperforms \cite{meier2008group} even if we do not manually engineer interactions among nucleotides.

Down-sampling or up-sampling is usually used for unbalanced data in the group lasso and the GLLR. However, we have the significant advantage of using the entire unbalanced training set because our SGIN is flexible enough to accommodate weighted cross entropy, and indeed any kind of loss function.
For fairness, we use the same set of hyper-parameters (neural network structure, number of epochs, learning rate, etc) when we compare our model to a down-sampling approach.

As shown in Table \ref{tab:rna_rst}, SGIN with weighted-loss outperforms \cite{yeo2004maximum,meier2008group} and 
finds that the first nucleotide postion in the 7-length sequence provides the least information for this binary classification task.
In fact, SGIN can achieve similar performance even without the first nucleotide position. 
Of note, our result suggests the first nucleotide position provides less information than the last nucleotide position because we already randomly permute the order of groups in Algorithm \ref{algo:sbcgd}; 
implying the importance of RNA sequence is not symmetric for this problem.

The lasso only performs slightly worse than the other models. 
When the number of sparse groups is larger than 4, the lasso performs similarly to our SGIN because the information left in the remaining variables is inadequate.

\subsection{Reduce Eye-Tracking Trials used to Discriminate between ASD and Non-ASD}
\label{sec:experiment_asd}

In the RNA splicing experiment,  the number of groups ($k = 7$) is small, and it is tractable to try a brute-force approach in which $O(2^k)$ models are used to test every combination of group inclusion/exclusion. Here, we use a more challenging dataset from an eye-tracking study for children with ASD, which contains $k = 109$ groups of variables. 

\begin{figure}[!h]
    \centering
    \begin{subfigure}[b]{0.32\textwidth}
        \includegraphics[width=0.99\textwidth]{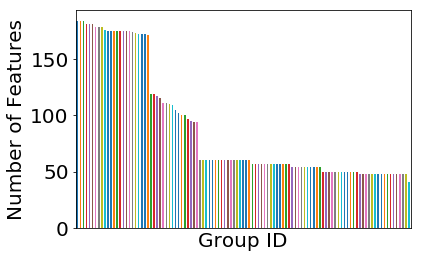}
        \caption{Num features in groups}
        \label{fig:asd_group_length}
    \end{subfigure}
    \begin{subfigure}[b]{0.32\textwidth}
        \includegraphics[width=0.99\textwidth]{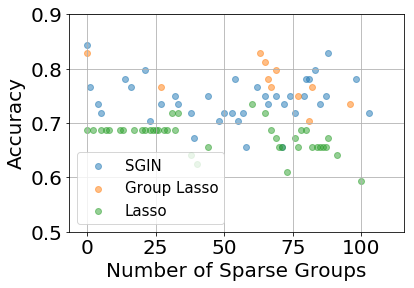}
        \caption{Validation Accuracy}
    \end{subfigure}
    \begin{subfigure}[b]{0.32\textwidth}
        \includegraphics[width=0.99\textwidth]{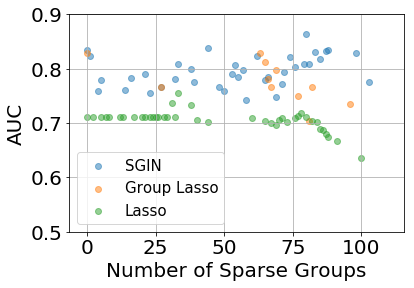}
        \caption{Validation AUC}
    \end{subfigure}
  \caption{ASD/non-ASD Classification Result for the Cross Validation. For comparison,
 a linear regression or logistic regression can achieve 81.25\% accuracy and 82.91\% AUC score in the validation set.
}
  \label{fig:asd}
\end{figure}

We recruited 64 children for an eye-tracking experiment that contains 109 stimuli designed by psychologists and clinicians to characterize known social impairments in ASD. 
In the post-hoc analysis, 9647 features from 109 groups are extracted, where each group corresponds to one eye-tracking stimulus. As shown in Figure \ref{fig:asd_group_length}, most groups have about 50 features, and 34 groups have more than 100 features.
Thirty-two participants were diagnosed with ASD, and the other thirty-two participants have different diagnostic labels (non-ASD). 
Missing values due to equipment failure or noise are filled  using the Expectation Maximization algorithm.

We apply eye tracking and area-of-interest analysis \cite{chawarska2012context} to the dataset to obtain clean features.
The eye-tracking stimuli and data cleaning pipelines were designed so that the extracted features can linearly separate the ASD/non-ASD group for monitoring and tracking of intervention effects in children with ASD.
Here, we test whether we need all 109 stimuli to achieve similar classification performance. 
As discussed, shorter experiments are more tolerable and can reduce participant burden for future eye-tracking studies. 
This is a $p >> k > n$ problem, where the number of features ($p=9647$) is much larger than the number of samples ($n=64$), and the number of groups ($k=109$) is also larger than the number of samples. Because of the large group size, we expect the machine learning models can learn to remove most of the groups.

We use 10-fold cross validation to validate the machine learning models. 
We acknowledge that the validation performance is over-optimistic and does not represent the real-world performance for ASD/non-ASD classification. 
ASD is a diverse and complex disorder that is diagnosed behaviorally by clinicians, a proccess that requires significant effort and professional training.
We use the same training/validation split for all models in the experiment so that the comparisons between different machine learning methods are fair.
We use the average number of sparse groups in the 10-fold cross validations as the result for one regularization term $\lambda$, as shown in Figure \ref{fig:asd}.

We use accuracy and area under the ROC curve (AUC) score to evaluate classification results.
Our result suggests even when 89.9\%  of the groups are removed (i.e., 99 stimuli are removed), SGIN can still achieve a satisfying classification result. 
The lasso is able to find group sparsity, but its performance is about 10\% worse than SGIN. 
The group lasso performs similarly to SGIN, though it struggles to learn fewer than 60 groups, even after tuning $\lambda$, which makes it difficult to reason about the least important groups of features: that is, the group lasso either removes no groups or many, and rarely anything in-between.
In the ablation study, the NN and the ``NN + $\tau$" perform slightly worse than SGIN even when the number of sparse groups is zero, and neither of them can find redundant groups for this problem.

Even with only 11 groups (out of the original 109 groups) of variables, SGIN still achieves 78.13\% accuracy and 82.91\% AUC score on the validation set.
This result suggests that substantial redundancy exists in the 109 stimuli: even using a small subset of these stimuli,  the eye-tracking experiment can still be effective in between-group classification.
Though these methods remain to be validated on larger, unseen test cases, our results suggest SGIN has the potential to help clinicians and psychologists design and select better and more streamlined eye-tracking stimuli in the future.


\subsection{Remove Rows for MNIST}
\label{sec:experiment_mnist}

We conduct 10-class classification experiments with the MNIST dataset \cite{lecun1995learning}, a classic dataset for  hand-written digit recognition, that, unlike the aforementioned datasets, is not linearly separable. Our goal is not to improve classification accuracy, which has been pushed over 99\% by existing methods \cite{wan2013regularization}. Instead, we are interested in identifying the rows in the MNIST images that are not necessary to recognize digits. 

The MNIST dataset contains 60000 images (samples) in its training set and 10000 images in its testing set. The size for each image is  $28 \times 28$ pixels.
In this experiment, we treat each row as a single group, and hence each image contains 784 features (pixels) in 28 groups.
We do not apply convolutional layers either to stay consistent with our other experiments.
We hold out 1000 samples from the training set for validation, and use accuracy to evaluate the testing result, as is convention.

\begin{figure}[!h]
    \centering
    \begin{subfigure}[c]{0.17\textwidth}
        \centering
        \includegraphics[width=0.45\textwidth]{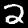}
        \includegraphics[width=0.45\textwidth]{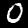}\\
        \includegraphics[width=0.45\textwidth]{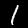}
        \includegraphics[width=0.45\textwidth]{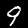}
        \caption{Raw}
        \label{fig:mnist_raw}
    \end{subfigure}%
    \begin{subfigure}[c]{0.17\textwidth}
        \centering
        \includegraphics[width=0.45\textwidth]{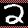}
        \includegraphics[width=0.45\textwidth]{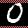}\\
        \includegraphics[width=0.45\textwidth]{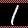}
        \includegraphics[width=0.45\textwidth]{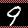}
        \caption{SG = 5}
    \end{subfigure}%
    \begin{subfigure}[c]{0.17\textwidth}
        \centering
        \includegraphics[width=0.45\textwidth]{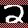}
        \includegraphics[width=0.45\textwidth]{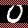}\\
        \includegraphics[width=0.45\textwidth]{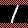}
        \includegraphics[width=0.45\textwidth]{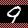}
        \caption{SG = 10}
    \end{subfigure}%
    \begin{subfigure}[c]{0.17\textwidth}
        \centering
        \includegraphics[width=0.45\textwidth]{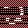}
        \includegraphics[width=0.45\textwidth]{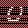}\\
        \includegraphics[width=0.45\textwidth]{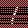}
        \includegraphics[width=0.45\textwidth]{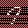}
        \caption{SG = 20}
        \label{fig:mnist_20}
    \end{subfigure}%
    \begin{subfigure}[c]{0.30\textwidth}
        \centering
        \includegraphics[width=0.99\textwidth]{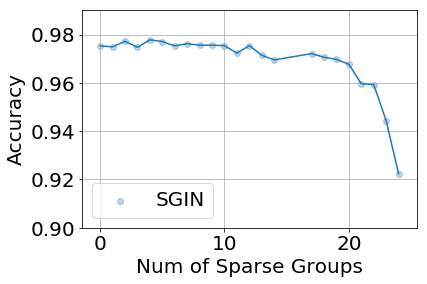}
    \end{subfigure}%
    \caption{
    Testing results for the MNIST Experiment:
    \textit{Left (\ref{fig:mnist_raw} - \ref{fig:mnist_20}):} 
    Red dotted rows have been made sparse by SGINs, where the number of sparse groups (SG) is tagged in captions.
    \textit{Right:} Number of sparse groups v.s. testing accuracy.
    Because we use a shallow model for experiments, the best accuracy is 98\% for the NN with no sparse rows. When SG = 10, SGIN can still achieve 97.65\% accuracy. When SG = 20, the accuracy is 94.33\%.
    }
    \label{fig:mnist}
\end{figure}

SGIN identifies unnecessary groups for images (rows) in the MNIST dataset 
without any human knowledge. When regularization term $\lambda$ 
is large for SGIN, the model will first remove the top and bottom rows, because these rows are usually black borders and do not contain any useful information. 
After approximately 10 rows from the top and 
bottom are made sparse, SGIN begins to remove every other row in the middle, since two consecutive rows usually provide similar information, and choosing separated rows maintains most of the necessary information. 
This is similar to a human’s ability to perceive an entire image while looking through window shutters (video interlacing effect), and it is also consistent with the intuition of pooling layers in convolutional neural network. 
Note that Lasso does not show interlacing, group lasso showed < 70\% testing accuracy, and more details about these linear methods are in the Appendix.

Surprisingly, we can still achieve satisfying performance (94.33\% testing accuracy) even if we only  use 8 groups (rows) of pixels out of the 28 groups. 
This result suggests SGIN has the potential to reduce data collection cost (e.g. purchasing less expensive cameras) for digit recognition  problems in the future. 
When all rows are used for digit recognition, the ``N + $\tau$" model performs 
slightly better than the NN and SGIN, but it could not find group sparsity in the  dataset. 
It only occasionally found one sparse group by chance even after a hundred  different $\lambda$  for regularization terms. 

To make the MNIST experiment more realistic, we conduct another experiment with Haar Wavelet Transformation, a common image compression method, to pre-process the dataset so that each data sample has 799 features in 13 groups. The SGIN can automatically learn that high frequency features and groups are not necessary for this classification task, and it can also detect some redundant lower frequency groups while $\lambda$ becomes large enough. Our result shows that only using 4 groups (40 features out of the original 784) can achieve 93\% testing accuracy, which is better than simply throwing out higher frequency features or down-sampling the image. This method could potentially be applied to efficiently stream data from IoT devices to cloud computing. The details and visualization for this wavelet experiment is in the Appendix.

\subsection{Stability}
This work aims to reduce data collection cost in inference time, and it is important to consider stability of group selection. 
Unfortunately, the stability cannot be theoretically guaranteed for multiple runs, because the column rank for $X$ is not full in all our experiments, which means the optimal solutions are not unique. 
We use mean pairwise Jaccard similarity (aka intersection over union) to measure the stability of groups for all experiments with 100 bootstrap runs.
 
In the eye tracking experiment, the stabilities for SGIN (0.73), lasso (0.78), and GLLR (0.82) are comparable. Lasso and group lasso are more stable than the SGIN in the MNIST experiment, even if SGIN has the best performance.
However, consider the following scenario for the MNIST experiment: a model sparsifies the even rows (row 2, 4, 6, ..., 28) in one run, and it sparsifies the odd rows (row 1, 3, 5, ..., 27) in another run; all popular stability measurements (including Pearson, Spearman, Tanimoto, etc) will indicate these two runs are perfectly unstable, but both runs provide reasonable sparsity decisions. Future studies can focus more on creating better measurements for stability.


\section{Discussion and Future Work}
\label{sec:discussion}

In all the experiments above, SGINs achieve the top or comparable-to-the-top performance. 
By using multi-layer non-linear neural network structure as the base inference model, SGIN can achieve high performance while excluding redundant groups. 
Even if the training data is linearly distributed, SGIN still has advantages over the lasso, the group lasso, and the GLLR. 
These experiments also show the optimization Algorithm \ref{algo:sbcgd} is essential for SGIN because the NN and the ``NN + $\tau$" models could not learn the group sparsity constraints with traditional SGD. 

When considering the eye-tracking experiment to distinguish ASD/non-ASD groups,
it is important to note that selecting a particular feature for a stimulus in an eye-tracking experiment is considered pivotal to psychologists, and organizing the stimuli into smaller sub-groups can help clinicians and researchers to study these potential biomarkers more efficiently. 
These explorations require sparsity in bi-level, hierarchical, tree-structure, and overlapping groups.
By adding another $l$-1 regularization term at the variable level, our model has the potential to detect bi-level sparsity, but the optimization algorithm needs another layer of iteration for variable-level optimization to consider individual weights. 
Similarly, $l$-2 regularization could be added to form a lasso-ridge bridge or elastic-net regularization for further regularization if necessary.
Reusing information across epochs is another possible idea to reduce training cost.
In the future, more studies should be conducted to extend SGIN to other various settings in real-world problems.

SGIN considers every group of input variables to have the same cost, but the cost to obtain features from different modalities is usually different in a real-world scenarios.
We could add a constant term $c_i$ into Equation \ref{eq:gsnn_reg} and get
 $\tau = \sum_{i=1}^k c_i \tau_i$, but more studies are needed to explore these convexity and optimization procedures.

SGIN system assumes every sample has complete data for all groups, but missing data is a common issue in reality for data processing and machine learning. 
In the future, we can create a more general model that can handle data with missing variables and groups.

\section{Conclusion}
\label{sec:conclusion}
In this study, we create a new loss function to regularize input groups in neural networks, design an effective optimization algorithm for the new loss function, and conduct three experiments to test the new loss and optimization algorithm. 
The results of our experiments suggest that SGIN can achieve sparsity in grouped input variables for neural networks while maintaining high accuracy. 
In particular, SGIN is able to produce satisfying results with one nucleotide position removed in the RNA splicing problem, with 89.9\% of the eye-tracking stimuli removed from the ASD/non-ASD classification problem, and with 20 rows removed from the hand-written digit recognition problem.
Future work will explore extensions of SGIN to additional architectures and problems.

\section*{Acknowledgement}
This work is supported by NIH awards K01 MH104739, R21 MH103550; the NSF Expedition in Socially Assistive Robotics \#1139078; and Simons Award \#383661.
We would also like to thank the anonymous reviewers for their helpful comments and suggestions.

\small

\bibliographystyle{unsrt} 
\bibliography{reference.bib}
\raggedbottom 

\input{appendix.tex}

\end{document}

%% file: appendix.tex
\pagebreak[4]

\appendix
\section*{Appendix}
\renewcommand{\thesubsection}{\Alph{subsection}}

\subsection{Method Details}

\subsubsection{Visualization of the Algorithm}
\begin{figure}[h]
    \includegraphics[width=0.9\textwidth]{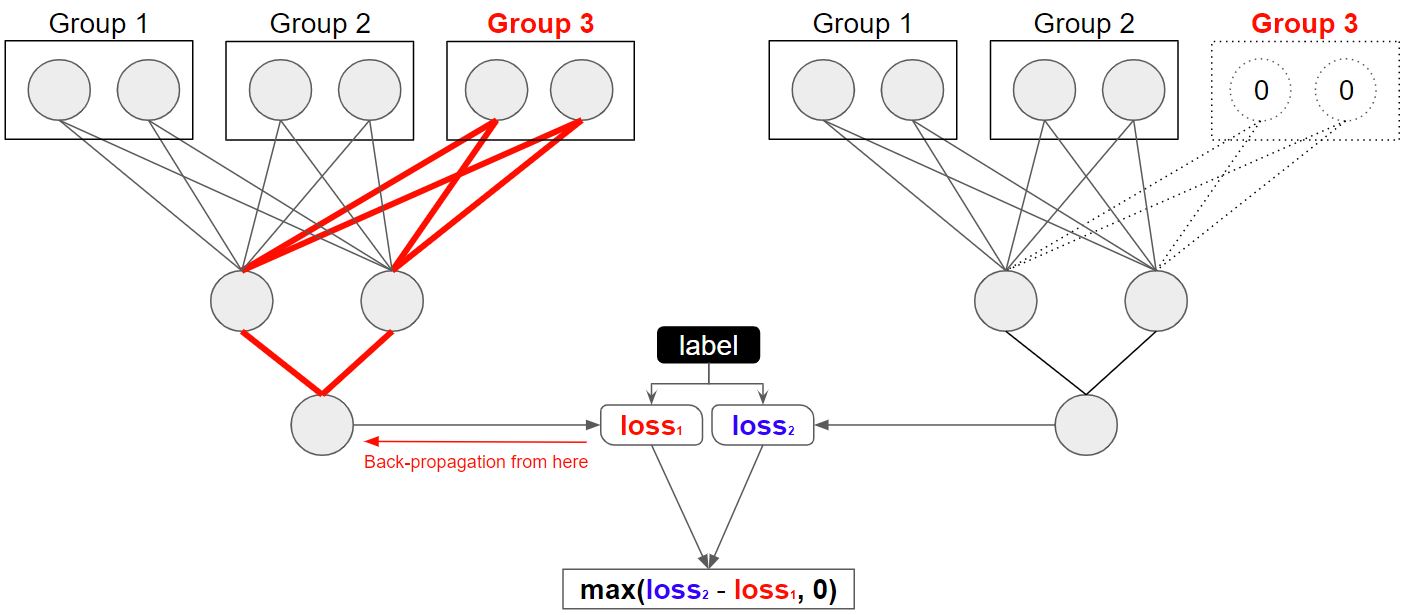}
    \caption{
    Illustration for line \ref{alg:line_loss_1} - line \ref{alg:line_sgd} for Algorithm \ref{algo:sbcgd}: for the NN, the first row is input layer, and the second row is the hidden layer. Suppose $p = 6$, $k = 3$, $p_i = 2$ for $\forall i$, we use one hidden layer with 2 neurons in the NN. We want to optimize the third input group (i.e. $i=3$). The left diagram shows how to calculate $loss_1$; right diagram shows how to calculate $loss_2$, where dashed lines represent disconnected parameters. The red bold lines in the left diagram show the back-propagation path for $loss_1 + \lambda \tau_i$. The bottom $max(loss_2 - loss_1, 0)$ is stored to compute coordinal loss for group 3 in current epoch.
}
\end{figure}

\subsubsection{Proofs}

\textbf{Proof for Proposition in Equation (\ref{eq:proposition}):}
\begin{align*}
&  & \mathbb{E}[\phi(f_{\beta_{\bar{i}}}(X_{\bar{i}}), y)]  -   \mathbb{E}[\phi(f_\beta(X), y)] < \lambda \tau_i\\
&  \Longleftrightarrow & \mathbb{E}[\phi(f_{\beta_{\bar{i}}}(X_{\bar{i}}), y)] < \mathbb{E}[\phi(f_\beta(X), y)] + \lambda \tau_i\\
&  \Longleftrightarrow & \mathbb{E}[\phi(f_{\beta_{\bar{i}}}(X_{\bar{i}}), y)] + \lambda \sum\limits_{j \neq i}^k \tau_j < \mathbb{E}[\phi(f_\beta(X), y)] + \lambda \sum\limits_{j=1}^k \tau_j\\
&  \Longleftrightarrow &  J( \beta_{\bar{i}}) < J(\beta)\\
\end{align*}

\subsubsection{Parity Simulation Experiment}
We conduct a parity experiment with noise to show the SGIN can learn interactions among different groups even if it optimizes the parameters for each group (in the first hidden layer) separately.
In this experiment, the input is a 10-dimensional array, where the 10 features are in 5 groups, and one of these groups (with 2 features) are just noise. This parity experiment can also be considered as an 8-way XOR operation.
We train a SGIN model to this operation with one-hidden layer with 20 neurons, and we use $\lambda = 10^{-4}$ as the regularization term.
We treat this problem as a binary classification problem,  assign 2 neurons in the output layer, randomly generate 10000 samples for training, decrease learning rate by 1\% after every epoch, and use cross entropy to train this model. Even with only one hidden layer, the SGIN can still learn interactions among all groups.  

\begin{figure}[h]
    \centering
    \begin{subfigure}[c]{0.59\textwidth}
        \includegraphics[width=0.99\textwidth]{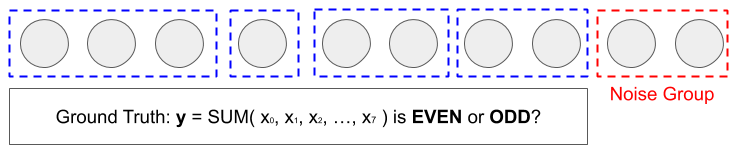}
        \caption{Simulation Setup}
    \end{subfigure}
    \begin{subfigure}[c]{0.40\textwidth}
        \includegraphics[width=0.99\textwidth]{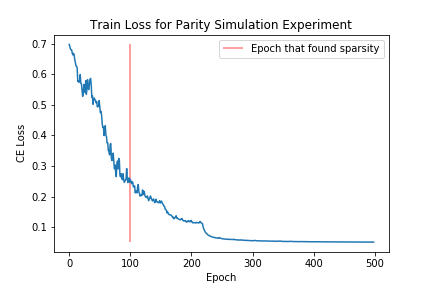}
        \caption{Training Loss}
    \end{subfigure}
    \caption{\textbf{The Partiy Simulation Experiment}: The SGIN model is able to detect the noise group near the 100th epoch, and learn the interactions among all other groups.}
\end{figure}

\subsubsection{A Variant of the SGIN to Guarantee Convergence}
Line \ref{alg:line_mean_arr} in Algorithm \ref{algo:sbcgd} use a heuristic way to calculate the expected difference among the two losses, where the losses are stored in an array during the mini-batch optimization.
A safer variant of the SGIN method is to directly compare the two losses before line \ref{alg:line_mean_arr}.  
This variant requires more training time for the SGIN model, but only has similar performance to the SGIN algorithm in the real-world scenario. 
Here, we formally present this algorithm. Algorithm \ref{algo:sbcgd_vb} runs 1.5 times slower than Algorithm \ref{algo:sbcgd}, but their final prediction accuracy are similar and within 1\% difference. 

\begin{algorithm}[h]
\SetAlgoLined
\LinesNumbered
\KwIn{Neural Network $f$, regularization term $\lambda$, loss function $\phi$, Training data $X$, training labels $y$}
\KwResult{Optimize the parameters $\mathbf{W}$ in the neural network $f$ for one epoch}
 Randomly shuffle the groups\;
\ForEach{Group $i \in [1,\ldots k]$}{
   \lIf{group $i$ is sparse in $f$}{ continue }
   arr = \{\}\;
   \ForEach{mini-batch $(X', y') \subset (X, y)$}{
       $loss = \phi(f(X'), y')$\;
       Optimize $\Theta \cup \theta_i$  according to $loss + \lambda \tau_i$  by using Stochastic Gradient Descent\;
    }
   \ForEach{mini-batch $(X', y') \subset (X, y)$}{
       $loss_1 = \phi(f(X'), y')$\;
       $loss_2 = \phi(f_{\bar{i}}(\overbar{X'_i}), y')$\;
       arr.push(max($loss_2 - loss_1$, 0))\;
   } 
   \If{mean(arr) $< \lambda$}{ 
       Set $\theta_i = 0$, and make group $i$ sparse in $f$\;
   }
}
\caption{Stochastic Blockwise Coordinated Gradient Descent (\textbf{Version B})}
    \label{algo:sbcgd_vb}
\end{algorithm}

\begin{figure}[!h]
    \centering
    \begin{subfigure}[b]{0.32\textwidth}
        \includegraphics[width=0.99\textwidth]{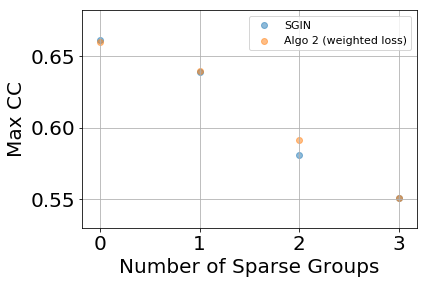}
        \caption{Nucleotides in RNA Splicing}
    \end{subfigure}
    \begin{subfigure}[b]{0.32\textwidth}
        \includegraphics[width=0.99\textwidth]{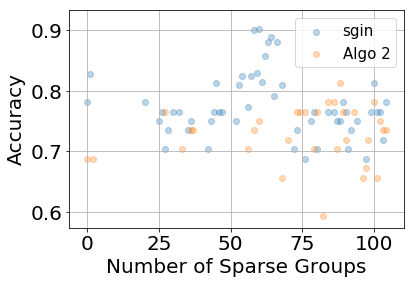}
        \caption{Reduce ET Trials}
    \end{subfigure}
    \begin{subfigure}[b]{0.32\textwidth}
        \includegraphics[width=0.99\textwidth]{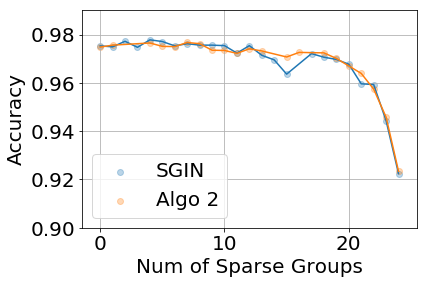}
        \caption{Remove Rows for MNIST}
    \end{subfigure}
  \caption{Results for Algorithm \ref{algo:sbcgd} and \ref{algo:sbcgd_vb}. The final performance are similar for both algorithms, but the SGIN (Algorithm \ref{algo:sbcgd}) is 50\% faster at training time.}
  \label{fig:version_b}
\end{figure}

\subsection{Engineering Details and Hyperparameter Searching}

\textbf{Implementation:} We use PyTorch package to implement the neural network (NN), SGIN, and optimization Algorithm \ref{algo:sbcgd}. In ablation studies, we use sklearn for the lasso, pyglmnet for the group lasso, and grplasso package provided by \cite{meier2008group} for the group lasso for logistic regression (GLLR). 
The grplasso package is used in R, and all other packages are used in Python 3.6.

\textbf{Experiment Data:} We use two public datasets in this study: the MNIST dataset is downloaded from python-mnist package (\url{https://pypi.org/project/python-mnist/}), and the RNA dataset is downloaded from (\url{http://hollywood.mit.edu/burgelab/maxent/ssdata/MEMset/}).

\textbf{Bias in First Hidden Layer:} we do not use bias in neurons in the first hidden layer for implementation convenience, but we use bias for all other layers.
\cite{scardapane2017group} also claims removing bias term would not affect prediction result.

\textbf{Non-linear Activation:} we use ReLU activation for all hidden layers in all experiments for consistency. We also test with Sigmoid activation, and the results are similar.

\textbf{Learning Rate:} we use the same initial learning rate ($\eta = 0.1$) for all the NN, the ``NN + $\tau$", and SGINs in all experiments.

\textbf{Batch Size:} we choose batch size purely based on computation time rather than prediction performance.
We use 100 as the batch size for RNA and ASD experiments, and 1000 for the MNIST experiment. 
Because the ASD dataset has less than 100 samples, Algorithm \ref{algo:sbcgd} does not perform stochastic fashion for this dataset and becomes Blockwise Coordinated Gradient Descent algorithm in this case. This example shows our loss function and optimiztion algorithm can converge even without the stochastic fashion for small datasets.

\textbf{Number of Epochs for SGIN:} because the complexity of the experiments are different, the number of training epochs are different in these experiments. We train SGINs 5 epochs for both the RNA splicing and the eye-tracking experiments, and we decrease the learning rate by 50\% after each epoch. For the MNIST experiment, we train SGINs for 10 epochs and decrease the learning rate by 10\% after each epoch. 
The number of epochs and learning rate decay are chosen based on convergence of training loss $\phi(f(X), y)$.

\textbf{Number of Epochs for the NN and the ``NN+$\tau$":} because Algorithm \ref{algo:sbcgd} loops at most $k$ times in line \ref{alg:line_loop_begin}; so, we train the NN and the ``NN + $\tau$" $k$ times more epochs than SGIN so that these models have similar number of back-propagation steps. This decision gives some slight disadvantage for SGIN, because once SGIN make a group sparse, it will reduce the number of back-propagation steps in future epochs. The learning decay rate is same as SGIN, but it will be applied after every $k$ epochs.

\textbf{Regularization Term:}
We test a large range of  $\lambda$ to ensure we can get different group sparsity for all models.
For the ``NN + $\lambda$" and SGIN, we use the same set of $\lambda$ terms, which ranges from $10^{-15}$ to $0.1$. For the lasso, the group lasso, and the GLLR, we test with $\lambda \in [10^{-15}, 10^5]$ depends on the dataset.

\textbf{Choice for Baseline Models:}
We use the GLLR for the RNA splicing dataset, because \cite{meier2008group} conducts a similar experiment. 
For the eye-tracking dataset, we only included the lasso and the group lasso for comparison because the GLLR performs slightly worse than the group lasso.
For the MNIST experiment, lasso, group lasso, logistic regression, $k$-nearest neighbour, and their variants do not perform as well as the NN, as discussed in \cite{lecun1995learning}. So, we exclude these models from the MNIST experiment. SGIN outperforms all the models mentioned above. Convolutional neural networks perform better than the NN and SGINs, but applying convolutional operations row-by-row to images is meaningless. More experiments should be conducted to apply our methods for convolutional layers in the future.

\textbf{Hyper-parameters for Baseline Models:} the only hyper-parameter we search for the lasso, the group lasso, and the GLLR is the $\lambda$ regularization term, and other hyper-parameters (e.g. stop criterion, maximum number of iterations, etc.) are the same for all experiments.

\subsection{Details for MNIST Experiments}

\begin{figure}
    \centering
    \begin{subfigure}[b]{0.49\textwidth}
        \includegraphics[width=0.99\textwidth]{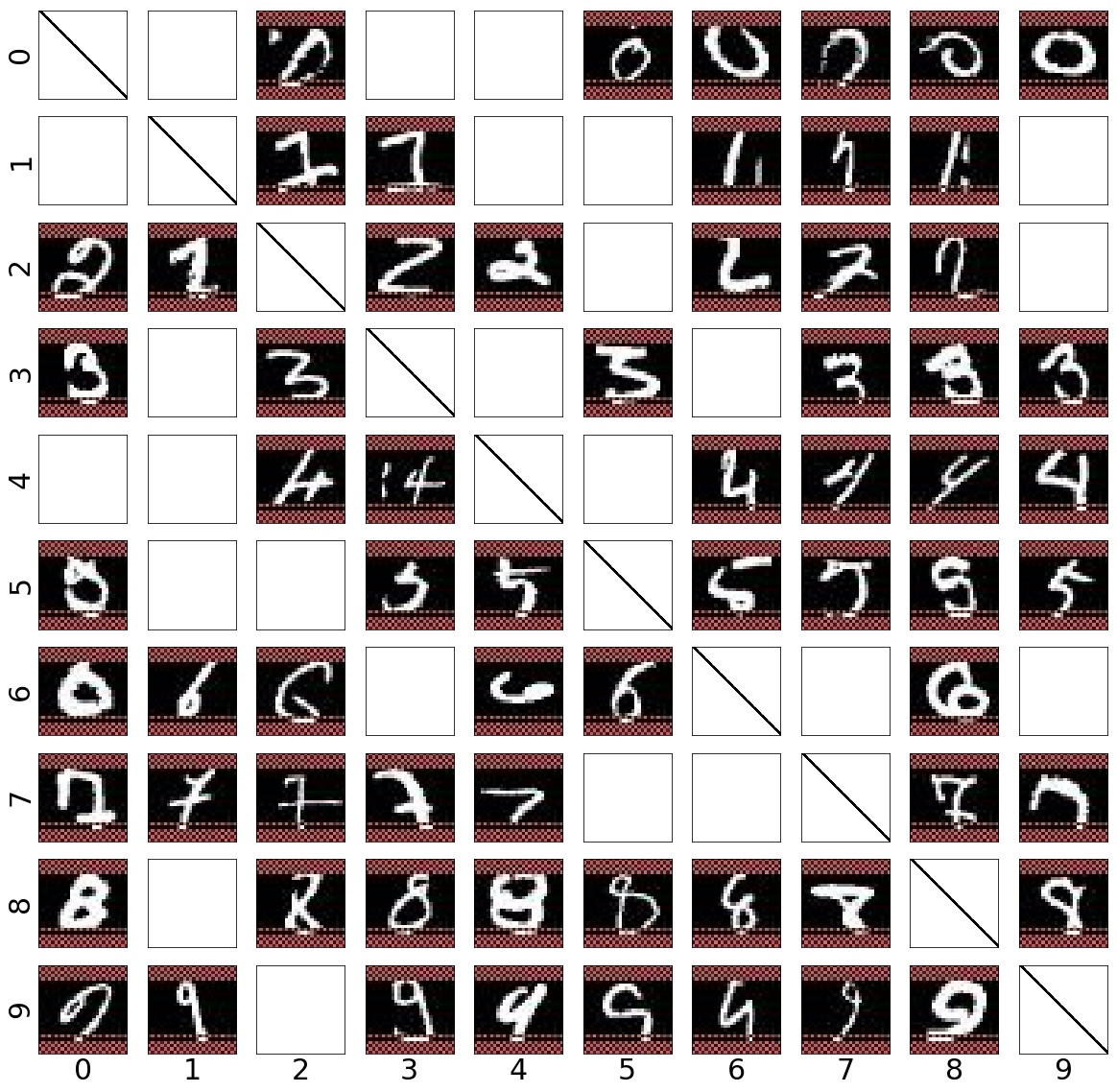}
        \caption{SG = 10}
    \end{subfigure}
    \begin{subfigure}[b]{0.49\textwidth}
        \includegraphics[width=0.99\textwidth]{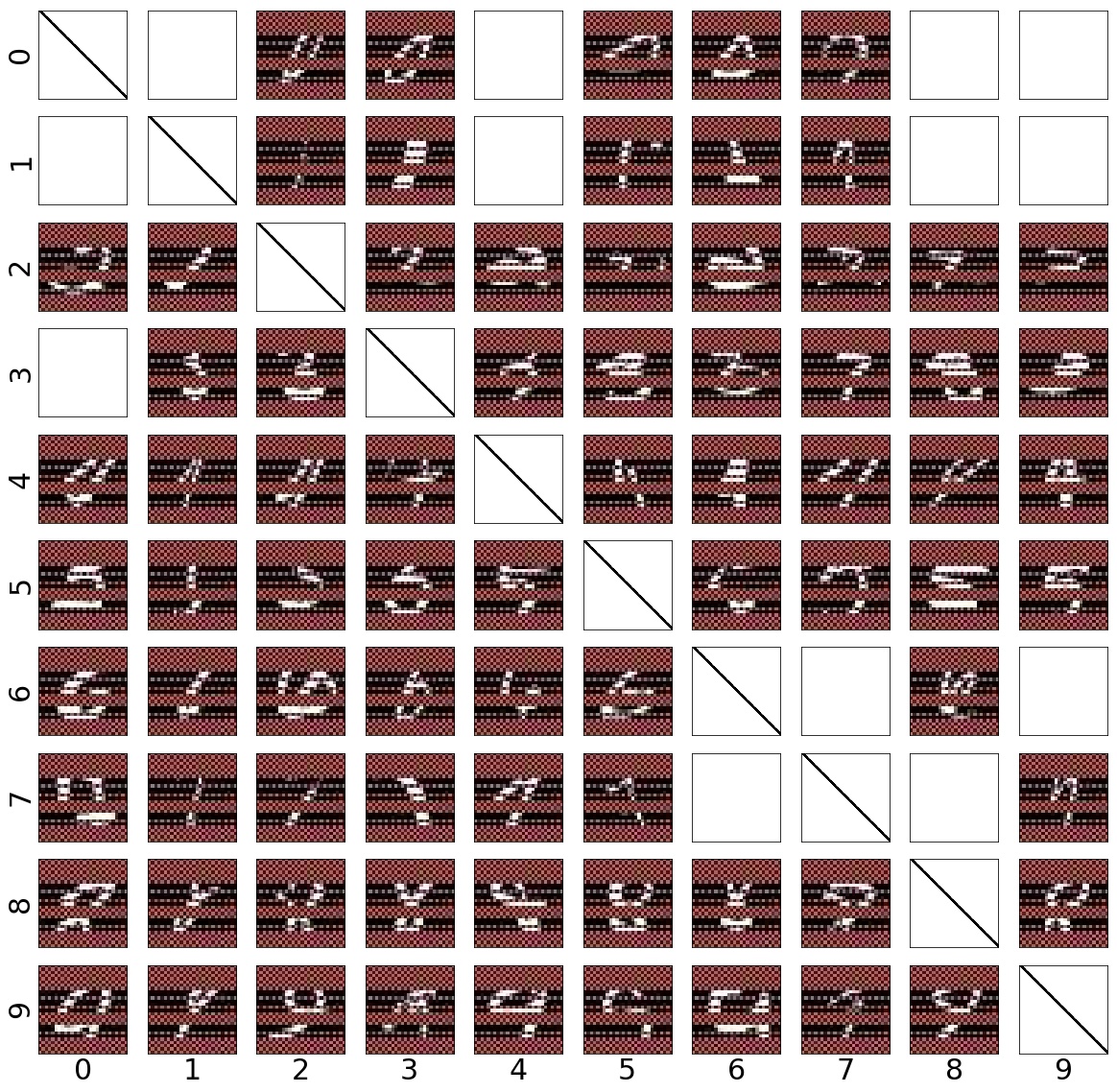}
        \caption{SG = 20}
    \end{subfigure}
    \caption{Confusion Matrix for the MNIST Experiment: in the matrix, the rows represent the ground truth, and the columns represent the prediction. We randomly select one image from the wrong predictions to show in the confusion matrix. White boxes represent no such (real, prediction) pair. The dotted red rows are sparsified and removed by SGIN.}
    \label{fig:mnist_cm}
\end{figure}

We provide two confusion matrices for the MNIST experiment in Figure \ref{fig:mnist_cm}. 
The sparse groups chosen by SGIN is different at each evaluation run, because we randomly shuffle the order of groups in Algorithm \ref{algo:sbcgd}. 

\subsubsection{Linear Models for the MNIST Row Experiment}
We use 1-vs-all method on linear models to perform the multi-class classification for the MNIST dataset. If a group is sparsified in all 10 sub-models, this group will be marked as ``sparse".

\subsubsection{MNIST Wavelet Experiment}

\begin{figure}[!h]
    \centering
    \begin{subfigure}[c]{0.17\textwidth}
        \centering
        \includegraphics[width=0.45\textwidth]{mnist_img/mnist_raw_id_47448.jpg}
        \includegraphics[width=0.45\textwidth]{mnist_img/mnist_raw_id_59256.jpg}\\
        \includegraphics[width=0.45\textwidth]{mnist_img/mnist_raw_id_42947.jpg}
        \includegraphics[width=0.45\textwidth]{mnist_img/mnist_raw_id_51981.jpg}
        \caption{Raw}
        \label{fig:mnist_raw_wavelet}
    \end{subfigure}%
    \begin{subfigure}[c]{0.17\textwidth}
        \centering
        \includegraphics[width=0.45\textwidth]{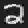}
        \includegraphics[width=0.45\textwidth]{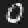}\\
        \includegraphics[width=0.45\textwidth]{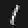}
        \includegraphics[width=0.45\textwidth]{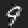}
        \caption{SG = 3}
    \end{subfigure}%
    \begin{subfigure}[c]{0.17\textwidth}
        \centering
        \includegraphics[width=0.45\textwidth]{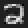}
        \includegraphics[width=0.45\textwidth]{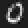}\\
        \includegraphics[width=0.45\textwidth]{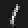}
        \includegraphics[width=0.45\textwidth]{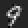}
        \caption{SG = 4}
    \end{subfigure}%
    \begin{subfigure}[c]{0.17\textwidth}
        \centering
        \includegraphics[width=0.45\textwidth]{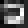}
        \includegraphics[width=0.45\textwidth]{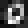}\\
        \includegraphics[width=0.45\textwidth]{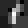}
        \includegraphics[width=0.45\textwidth]{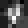}
        \caption{SG = 8}
        \label{fig:mnist_8_wavelet}
    \end{subfigure}%
    \begin{subfigure}[c]{0.30\textwidth}
        \centering
        \includegraphics[width=0.99\textwidth]{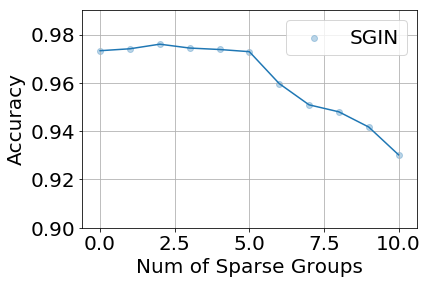}
    \end{subfigure}%
    \caption{
    Results for the MNIST Wavelet Experiment:
    \textit{Left (\ref{fig:mnist_raw_wavelet} - \ref{fig:mnist_8_wavelet}):} 
    Recovered image using Inverse Haar Wavelet Transform, where frequency-domain features in removed groups are set to zero.
    \textit{Right:} Number of sparse groups v.s. testing accuracy.
    }
    \label{fig:mnist_wavelet}
\end{figure}

The detailed results for the wavelet experiment for the MNIST experiment is shown in Figure \ref{fig:mnist_wavelet}. 
Because the width of the raw image is not divisible by 4, border effects create more features for each image in the 4-level Haar Wavelet Transform.
So, the wavelet transform created 799 features for each 28 x 28 image. 
The 799 features can be placed into 13 groups, where the first group contains $2x2=4$ features from the lowest level, and the remaining 12 groups are horizontal, vertical, and diagonal features for each of the 4 frequency levels. 
The size for each group depends on the level of the features: groups from the first level contains $2x2=4$ features, groups from the second level contains, $4x4=16$ features, groups from the third level contains $7x7=49$ features (border effects occur here), and groups from the fourth level contains $14x14=196$ features. As seem, groups in higher frequency levels contains more features but affect the image quality the less. 
So, high level features are usually discarded in image compression algorithms, such as JPEG.


\begin{figure}[!h]
    \centering
    \begin{subfigure}[b]{0.32\textwidth}
        \includegraphics[width=0.99\textwidth]{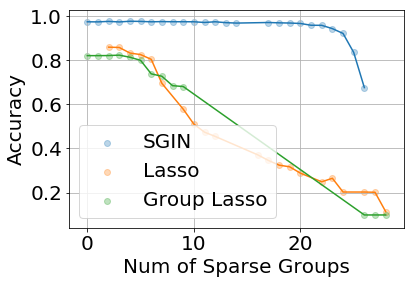}
        \caption{Row-wise Experiment}
        \label{fig:mnist_raw_linear}
    \end{subfigure}
    \begin{subfigure}[b]{0.32\textwidth}
        \includegraphics[width=0.99\textwidth]{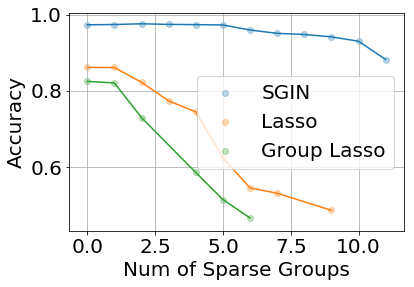}
        \caption{Wavelet Experiment}
        \label{fig:mnist_wavelets_linear}
    \end{subfigure}
  \caption{MNIST Experiment Results with Linear Methods}
\end{figure}

